\documentclass[10pt,twocolumn,letterpaper]{article}

\usepackage[pagenumbers]{cvpr} 

\usepackage{graphicx}
\usepackage{amsmath}
\usepackage{amssymb}
\usepackage{booktabs}
\usepackage{multirow}
\usepackage{multicol}
\usepackage{pifont}
\newcommand{\cmark}{\ding{51}}
\newcommand{\xmark}{\ding{55}}

\usepackage[pagebackref,breaklinks,colorlinks]{hyperref}

\usepackage[capitalize]{cleveref}
\crefname{section}{Sec.}{Secs.}
\Crefname{section}{Section}{Sections}
\Crefname{table}{Table}{Tables}
\crefname{table}{Tab.}{Tabs.}

\usepackage{arydshln}

\def\cvprPaperID{0011} 
\def\confName{CVPR}
\def\confYear{2023}

\begin{document}

\title{On Advantages of Mask-level Recognition for Outlier-aware Segmentation}

\author{Matej Grcić, Josip Šarić, Siniša Šegvić \\
University of Zagreb, Faculty of Electrical Engineering and Computing\\
Unska 3, 10000 Zagreb, Croatia\\
{\tt\small \{name.surname\}@fer.hr}
}
\maketitle

\begin{abstract}
Most dense recognition approaches
bring a separate decision
in each particular pixel.
These approaches 
deliver
competitive performance
in usual closed-set setups.
However, important 
applications in the wild
typically require 
strong performance in presence of outliers.
We show that 
this
demanding setup
greatly benefit 
from mask-level predictions,
even in the case of 
non-finetuned baseline models.
Moreover, we propose 
an alternative formulation
of dense recognition uncertainty
that effectively reduces
false positive responses 
at semantic borders.
The proposed formulation
produces a further improvement
over a very strong baseline
and sets the new state of the art
in outlier-aware semantic segmentation 
with and without training on negative data.
Our contributions also lead
to performance improvement
in a recent 
panoptic setup.
In-depth experiments confirm
that our approach succeeds
due to implicit aggregation
of pixel-level cues
into mask-level predictions.
\end{abstract}

\section{Introduction}
\label{sec:intro}

Emergence of deep learning 
revolutionized the field of computer vision \cite{krizhevsky12neurips}.
Complex yet efficient deep networks advanced the capability of machines to understand scenes \cite{farabet13pami,zbontar16jmlr}.
Segmentation is a very important form of scene understanding due to its applications in medicine, agriculture, robotics and the automotive industry.
In the last decade, segmentation tasks were modelled as per-pixel classification \cite{farabet13pami,long15cvpr}.
However, such approach assumes independence of neighbouring pixels, which does not hold in practice.
Neighbouring pixels are usually strongly correlated due to belonging to the same object or scene part \cite{lin14eccv}.
Albeit designed and trained with false assumption on independence of neighbouring pixels,
the obtained models deliver competitive generalization performance in in-distribution scenes \cite{chen18eccv,cheng20cvpr}.
However, their 
real-world performance 
still leaves much to be desired 
due to insufficient 
handling of the out-of-taxonomy scene parts  
\cite{blum19iccvw,chan21neuripsd}. 

\begin{figure*}[ht]
 \centering
 \includegraphics[width=\linewidth]{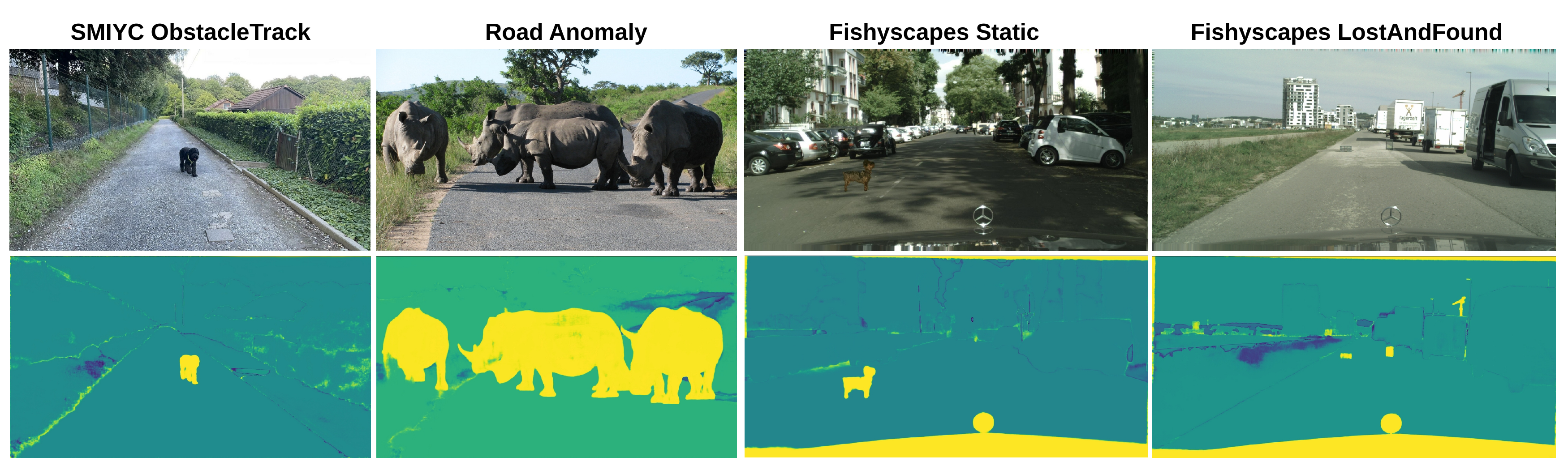}
 \caption{Outlier-aware segmentation with
   the proposed mask-level approach.
   We present input images (top)
   and dense OOD scores
   (bottom).
 }
 \label{fig:m2f-ad}
\end{figure*}

A recent approach to per-pixel classification decouples localization from recognition \cite{cheng21neurips}.
The localization is carried out 
by assigning pixels to an
abundant set of masks, 
each trained to capture 
semantically related regions 
(e.g.\ a road or a building).
The recovered semantic regions are subsequently classified as a whole.
The described approach is dubbed mask-level recognition  \cite{cheng22cvpr}.
Decoupling localization from classification further enables utilizing the same model for semantic, instance and panoptic segmentation.
The shared architecture performs competitively on standard segmentation benchmarks \cite{lin14eccv,cordts16cvpr,zhou19ijcv}.

However, prior work 
does not consider 
demanding applications of
mask-based approaches.
Thus, we investigate the value 
of mask-level recognition 
in some of the last 
major remaining challenges
towards scene understanding
in the wild - outlier-aware semantic segmentation \cite{blum21ijcv,chan21neuripsd,hwang21cvpr}
and outlier-aware panoptic segmentation \cite{hwang21cvpr}.
Our experiments reveal 
strong performance of 
mask-level approaches 
in these challenges.
We investigate 
the reasons behind such behaviour
and contribute improvements 
that support these important
applications.

Mask-level recognition has 
several interesting properties.
For instance, masks are classified
into K known classes 
and the class void,
while mask assignments 
are not mutually exclusive \cite{cheng21neurips}.
This provides 
more opportunity 
to reject predictions
than in standard 
per-pixel approaches.
Mask-level approaches 
can propagate mask-level uncertainty 
to the pixel-level. 
This is different 
from the standard approach 
which has to estimate 
independent anomaly scores in each pixel \cite{hendrycks22icml}. 
Obviously, the standard approach 
can easily ignore 
the local correlations
in a pixel neighborhood,
which does not
seem desirable. 
In terms of scalability, 
mask-level recognition models
do not require 
per-class feature maps
at the output resolution.  
This allows designers
to decrease the training footprint
\cite{bulo18cvpr}
and increase 
the flexibility of training.
All these properties 
make mask-level recognition
a compelling research topic.

This paper proposes 
the following contributions.
We point out that 
mask-level recognition
delivers strong 
baseline performance 
on standard benchmarks
for 
outlier-aware segmentation.
Our improvements 
further exploit
the specific bias 
of mask-level recognition.
Combining the proposed EAM outlier detector
with negative supervision
attains competitive results in 
outlier-aware semantic and panoptic segmentation.
Further improvements can be obtained 
by combining the proposed approach 
with negative supervision.
The resulting models 
set the new
state of the art 
in outlier-aware segmentation
on two tracks from the
Segment Me If You Can 
(SMIYC) benchmark and adapted MS COCO.

\section{Related work}
\label{sec:rw}

The related work considers models for mask-level recognition tasks (Sec.\ \ref{subsec:rw_mask_cls}) and segmentation in presence of outliers (Sec.\ \ref{subsec:rw_osr}).


\subsection{Recognition of free-form regions}
\label{subsec:rw_mask_cls}

Early approaches to mask-wide recognition relied on class-agnostic bottom-up proposals.
They aggregated hand-crafted \cite{carreira12eccv} or convolutional \cite{hariharan14eccv,dai15cvpr,pinheiro16eccv} features along the proposed regions and brought mask-wide decisions by classifying pooled representations.
Mask-RCNN extends this approach by sharing features across detection of proposals and mask-wide classification, as well as by end-to-end training of all parameters.
Recently, PointRend proposes to back-propagate the loss only through selected low-uncertainty predictions \cite{kirillov20cvpr}.
This allows to increase mask-RCNN resolution from 28$\times$28 to 224$\times$224 with a neglectable impact on the training footprint.
Very recently, MaskFormer precludes dependence on bottom-up proposals by directly assigning pixels to masks that span arbitrary image regions \cite{cheng21neurips}.
Its key component is a hypernetwork \cite{ha17iclr} that produces the weights for two 1$\times$1 convolutions that convert pixel-level embeddings into mask assignment scores and, subsequently, into semantic maps.
This is the first architecture that succeeds to deliver competitive experimental performance on three dense recognition tasks: semantic segmentation, instance segmentation, and panoptic segmentation.
Mask2Former \cite{cheng22cvpr} further improves the mask hypernetwork by introducing a special kind of attention layer that promotes progressive focusing onto foreground pixels for a particular mask.
Our work explores the Mask2Former performance 
in the context of outlier-aware
segmentation and outlier-aware panoptic segmentation.


\subsection{Segmentation in presence of outliers}
\label{subsec:rw_osr}
Recognition in the wild
involves test regions
beyond the training taxonomy.
Adequate models should reject the decision
in such pixels \cite{scheirer12tpami}.
This can be carried out by restricting
the shape of the decision boundary
\cite{scheirer14tpami,bendale16cvpr}
or by complementing the classifier
with an anomaly detector
\cite{hendrycks17iclr,liang18iclr}.
The decision boundary can be restricted
by thresholding distance
from the learned class centers
in the embedding space
\cite{scheirer14tpami,cen21iccv}.
This can be further improved by
employing a stronger classifier \cite{vaze22iclr}.
Nevertheless, many of these approaches 
are bound to fail if 
unknown samples happen to map 
to the same features as the samples
from the known classes.
This occurrence is known 
as feature collapse \cite{lucas19neurips}.

Early approaches for extending
discriminative predictions 
with OOD detection 
have been based on
prediction confidence \cite{hendrycks17iclr},
input perturbations \cite{liang18iclr},
density estimation \cite{nalisnick19iclr} and
Bayesian uncertainty \cite{mukhoti18arxiv}.
Several studies point out that
semantic anomalies \cite{ruff21pieee}
may be especially hard to detect
\cite{nalisnick19iclr,serra20iclr,kirichenko20neurips}.
A promising approach involves
generating synthetic anomalies
in tandem with the discriminative task
\cite{lee18iclr,zhang20eccv,grcic21visapp,chen22tpami}.
Further empirical improvements
have been achieved
by mimicking anomalies with
negative training data
\cite{hendrycks19iclr,liu20neurips}.
However, this may lead
to over-optimistic performance estimates
due to
possible
overlap with test anomalies.

Outlier detection is especially
interesting in the dense prediction context
due to important applications
in robust scene understanding
\cite{blum21ijcv,chan21neuripsd,zendel18eccv}.
However, straight-forward adaptations
of image-wide approaches
experience two important failure modes.
First, they often fail
to accurately localize anomalies
in front of inlier backgrounds
\cite{bevandic19gcpr}.
Second, they are prone to false positives
in inlier pixels with high entropy predictions
that occur regularly at semantic borders
\cite{rottmann20ijcnn}.
Hence, a large body of work proposes
custom designs
to alleviate these problems.

Partially anomalous images can be accounted for
by learning on mixed-content images
\cite{bevandic19gcpr,grcic22eccv,biase21cvpr,tian22eccv}.
Correlation between neighbouring pixels
can be addressed by aggregating evidence
through meta-classification \cite{rottmann20ijcnn}
or input pre-processing
\cite{liang18iclr}.
Real training data 
can be avoided 
by fitting generative heads
to pre-trained \cite{blum21ijcv}
or jointly trained \cite{hwang21cvpr,liang22neurips} features.
Another line of work trains on synthetic negatives
corresponding to adversarial noise \cite{besnier21iccv}
or samples of a jointly trained generative model \cite{grcic21arxiv}.
Finally, some approaches detect the discrepancy
between the input and the resynthesised scene
\cite{lis19iccv,xia20eccv,biase21cvpr,vojir21iccv}.

Different than all previous works,
we formulate 
outlier
detection
according to mask-wide predictions.
Different than meta-classification approaches
\cite{rottmann20ijcnn,chan21iccv}
our method requires only one learning episode
and does not require negative data.
Our method is orthogonal
to most previous approaches
and it, therefore, represents an exciting
baseline for future work.

\section{Mask-level recognition in presence of outliers}
\label{sec:method}

\begin{figure*}[h]
    \centering
    \includegraphics[width=\linewidth]{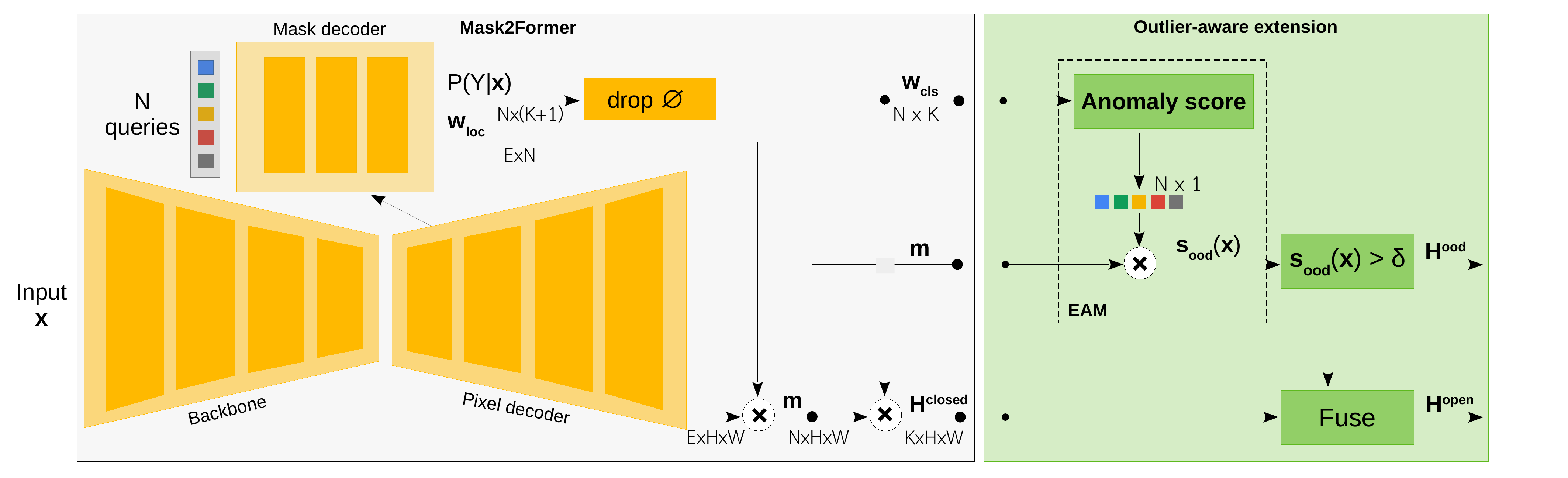}
 \caption{
  We focus on three tensors
  that are produced
  by the standard M2F model
  (left)
  \cite{cheng22cvpr}:
  closed-set segmentation 
  $\mathbf{H}_\mathrm{closed}$
  (K$\times$H$\times$W),
  per-mask dense binary assignments 
  $\mathbf{m}$ (N$\times$H$\times$W),
  and image-wide mask-level class scores 
  $\mathbf{w}_\mathrm{cls}$
  (N$\times$K).
  We start our outlier-aware extension
  (right)
  by quantifying uncertainty
  of mask-level predictions 
  $\mathbf{w}_\mathrm{cls}$.
  We recover the dense anomaly map
  $\mathbf{s}_\mathrm{OOD}^\mathrm{EAM}$
  (H$\times$W)
  by redistributing 
  per-mask anomaly scores 
  back to the pixels according to 
  dense mask assignment $\mathbf{m}$
  as shown in (\ref{eq:mask_cls}).
  We assemble outlier-aware segmentation 
  $\mathbf{H}_\mathrm{open}$
  by thresholding 
  $\mathbf{s}_\mathrm{OOD}^\mathrm{EAM}$
  and fusing it with 
  $\mathbf{H}_\mathrm{closed}$.  
  Note that $\sum_{rc}
   \mathbf{m}_i[r,c]\neq 1$.
 } 
 \label{fig:m2f-osr}
\end{figure*}

We present a novel approach
for extending mask-level dense prediction
towards outlier-aware segmentation.
Our approach 
can operate above 
many of the recent
dense prediction approaches
based on mask-level recognition 
\cite{hwang21cvpr,cheng21neurips,li2022arxiv}.
We formulate a novel 
dense OOD score
by ensembling mask-wide 
anomaly scores.
This improves outlier-aware segmentation 
on real datasets due to 
aggregating pixel-level evidence
across image regions
and decreasing sensitivity 
to semantic boundaries.

\subsection{Semantic segmentation with mask-level recognition}
\label{sec:review_m2f}

Mask-level segmentation approaches decouple
classification from localization 
and model them with 
separate prediction heads
\cite{cheng21neurips}.
Localization can be formulated 
through probabilistic 
assignments (masks)
$\mathcal{S} = \{\mathbf{m}_i \, | \, i=1,\dots,N\}$ 
that capture semantically related regions.
Each mask $\mathbf{m}_i$ 
is an $H\times W$ array 
of probabilistic assignments 
to the corresponding pixel.
We can join masks into 3D tensor $\mathbf{m}^{N\times H \times W}$.
Masks are recovered by subjecting  
standard dense features $\mathbf{E}$ 
to inferred projection $\mathbf{w}_\mathrm{loc}$ 
and sigmoid activation:
\begin{equation}
 \mathbf{m} = 
  \sigma(\mathrm{conv}_{1\times 1}
  (\mathbf{E},\mathbf{w}_\mathrm{loc})).
\end{equation}
Recognition can be carried out by inferring 
N mask-wide categorical distributions
into K known classes and one void class.
We denote these predictions as 
$P_i(Y=k|\mathbf{x})$,
$i\in$ 1..N, 
$k\in$ 1..K+1.
Let us consider probabilities of non-void classes
and arrange them into a $N\times K$ matrix 
$\mathbf{w}_\mathrm{cls}$.
Then the tensor of closed-set 
semantic segmentation scores
can be recovered by projecting masks 
according to $\mathbf{w}_\mathrm{cls}$:
\begin{equation}
 \mathbf{H}_\mathrm{closed} = 
  \mathrm{conv}_{1\times 1}
  (\mathbf{m}, \mathbf{w}_\mathrm{cls}).
\end{equation}
Note that this tensor 
does not contain distributions
since $\sum_i m_i[r,c]\neq 1$
and 
$\sum_k 
  \mathbf{w}_\mathrm{cls}[i,k]\neq 1$. 
The above convolution can be interpreted 
as classifying each pixel (r,c)
according to a weighted ensemble 
of per-mask classifiers
where the weights correspond 
to dense mask assignments:
\begin{equation}
  \hat{y}[r,c] = 
  \underset{k=1\dots K} 
  {\mathrm{argmax}}
  \sum_i \mathbf{m}_i[r,c]
    \cdot
    P_i(Y=k|\mathbf{x})
  \;.
  \label{eq:m2f-ens}
\end{equation}

%
Figure \ref{fig:m2f-osr} (left) shows 
that dense features $\mathbf{E}$
are produced in usual fashion, 
by connecting an off-the-shelf backbone
to an upsampling decoder 
with skip connections.
The main novelty is a hypernetwork
denoted as mask decoder
that receives latent features
and infers image-wide weights 
$\mathbf{w}_\mathrm{loc}$ 
and $\mathbf{w}_\mathrm{cls}$. 
The training fits
mask assignments $\mathbf{m}$
and mask-level recognition 
$P_i(Y=k|\mathbf{x})$ 
to the dense labels.

\subsection{Detecting outliers in pixel-level predictions}
\label{sec:pixel_ad}

Dense OOD detection 
requires a scoring function 
$\mathbf{s}_\mathrm{ood}: 
  [0,255]^{3 \times H \times W}
  \rightarrow 
  \mathcal{R}^{H\times W}$ 
that maps each pixel 
to the corresponding anomaly score.
Subsequently, we can 
detect anomalies by thresholding
the anomaly score 
$\mathbf{s}_\mathrm{ood}
(\mathbf{x})$. 
We can recover 
outlier-aware segmentation
by fusing anomalies with 
closed-set segmentation.

Several standard baselines
detect anomalous regions
according to uncertainty
of pixel-level predictions
\cite{blum21ijcv,hendrycks22icml}.
The prediction uncertainty 
can be quantified as
max-score \cite{hendrycks17iclr}, 
entropy \cite{chan21iccv}, 
energy \cite{liu20neurips} etc.
We shall evaluate that approach
by the PerPixel baseline that ablates
the mask decoder
and replaces it with standard 
per-pixel predictions 
\cite{long15cvpr}.

\begin{figure}[ht]
 \centering
 \includegraphics[width=\linewidth]
 {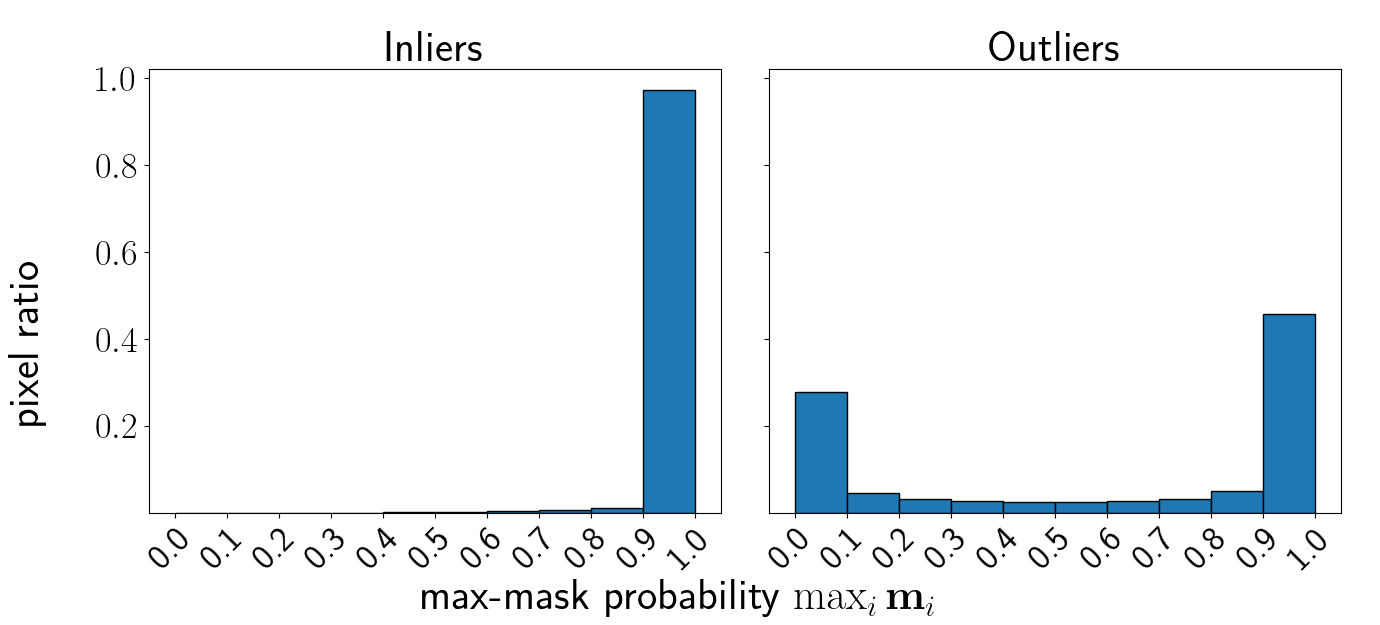}
 \caption{
 	Relative pixel frequencies according 
 	to max mask probability
 	in inlier and outlier pixels
 	on Fishyscapes L{\&}F val.}
 \label{fig:laf_max_sigma_hist}
\end{figure}

Pixel-level predictions
can also 
be recovered
with a mask-level model.
The training procedure
encourages masks 
$\mathbf{m}_i$
to specialize for capturing 
specific visual concepts.
Hence, one could define 
a pixel-level anomaly score 
which rejects pixels 
that are not claimed 
by any mask:
\begin{equation}
\label{eq:max_sigma}
 \mathbf{s}_\mathrm{ood}^\mathrm{AM}
  (\mathbf{x})[r,c] =  
  -\max_{i} 
    \mathbf{m}_i[r,c]
\end{equation}
AM stands for Anomaly 
of the max-Mask.
Accordingly, we shall have 
a high anomaly score where all masks
have low confidence.
Even though this approach 
outperforms the per-pixel baseline,
it is far from perfect. 
Fig.\ \ref{fig:laf_max_sigma_hist}
shows histograms of
inliers and outliers 
on Fishyscapes L{\&}F val
according to $\max{m}_i$ score.
The left histogram reveals 
that almost all inliers 
have high-confidence 
mask assignments.
On the other hand, 
the outlier distribution 
is highly polarized.
The left mode can be easily
distinguished from inliers,
but the right mode 
presents a tougher challenge.
This suggests that 
pixel-level predictions
may not be an optimal
solution to our problem,
because many of the 
real outlier pixels
get high confidence
mask assignments.
Therefore, we consider to build 
on mask-level uncertainty.

\subsection{Detecting outliers in mask-level predictions}

We first consider a method that
recovers dense anomaly scores
as mask-level uncertainty 
of the strongest mask.
If we choose max-softmax 
as the uncertainty measure,
we can formulate 
this score as:
\begin{equation}
\label{eq:s_hw0}
 \mathbf{s}
   _\mathrm{ood}
   ^\mathrm{AHM}
  (\mathbf{x})[r,c] =  
  - \!\! \max_{k=1\dots K} 
    P_{\mathrm{argmax}_i \mathbf{m}_i[r,c]} (Y=k | \mathbf{x}).
\end{equation}
AHM stands for
Anomaly score
of Hard-assigned Masks.
However, this approach 
completely ignores
the uncertainty of
the dominant mask assignment.
This clearly feels suboptimal
and our empirical results
confirm this intuition.
Therefore, we set out 
to combine uncertainties 
of pixel-level mask assignment 
and mask-level recognition.

We proceed by considering 
closed-set semantic 
segmentation scores (\ref{eq:m2f-ens}).
We can quantify their uncertainty
according to an arbitrary anomaly detector.
If we choose max-logit detector
\cite{hendrycks22icml},
we obtain the following:
\begin{equation}
\label{eq:s_hw}
 \mathbf{s}
  _\mathrm{ood}
  ^\mathrm{AEM}
  (\mathbf{x})[r,c] =  
  - \!\! \max_{k=1\dots K} 
   \sum_i
    \mathbf{m}_i[r,c]
    \cdot 
    P_i(Y=k | \mathbf{x}).
\end{equation}
Closed-set semantic scores
can be viewed as ensembled
outputs of per-mask classifiers,
where mask assignments act as
weights of the ensemble members.
Hence, we denote this score as
Anomaly of Ensembled 
Mask-wide predictions (AEM).

Finally, we consider 
to apply anomaly detector
directly to mask-level
classification scores.
We propose to aggregate
the resulting evidence 
in each particular pixel 
according to its 
mask assignments 
$\mathbf{m}$.
This approach can be interpreted
as an Ensemble over
Anomaly scores of
Mask-wide predictions (EAM).
This approach has an 
intuitive appeal
due to direct relation towards
mask-level uncertainty.
If we quantify 
mask-level uncertainty 
according to 
maximum per-class probability,
we get a lower bound 
of the AEM score (\ref{eq:s_hw}):
\begin{multline}
\label{eq:mask_cls}
 \mathbf{s}_\mathrm{ood}^{\mathrm{EAM}}
  (\mathbf{x})[r,c] 
  =  
  \sum_{i} \mathbf{m}_i[r,c] 
  \cdot  
  (-\max_{k=1\dots K}
   P_i(Y=k | \mathbf{x})) \\
  \leq - \max_{k=1\dots K} 
  \sum_{i} \mathbf{m}_i[r,c] 
  \cdot 
  P_i(Y=k | \mathbf{x})
\end{multline}
Fig.\ \ref{fig:m2f-osr}
(right)
illustrates steps to 
compute the EAM score
from M2F outputs.

\begin{figure}[ht]
 \centering
 \includegraphics[width=\linewidth]
  {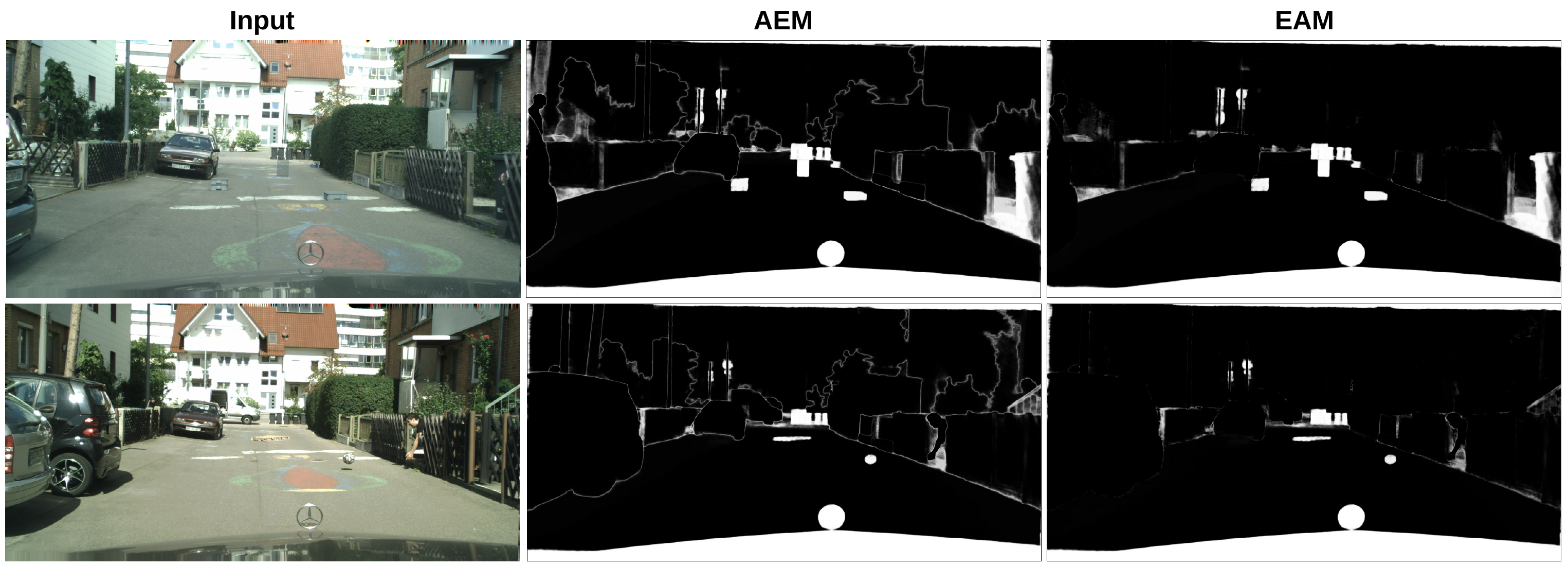}
 \caption{
 Pixel-level vs.\ mask-level OOD detection. Mask-level OOD detection alleviates the known issue of false positives at semantic borders. Please zoom in for the details.
 }
 \label{fig:heatmap}
\end{figure}

We expect that the difference
between the two approaches
should be best visible
at semantic borders.
Here adjacent masks
often lower their 
pixel assignment confidence.
In such situations
our proposed EAM approach
will correctly output
a lower anomaly score than AEM.
Fig.\ \ref{fig:heatmap}
illustrates the differences
between EAM and AEM scoring
on two scenes from
Fishyscapes L{\&}F.
We observe a similar behaviour 
in most of image pixels.
However, the proposed EAM approach
clearly outputs
lower anomaly score
on semantic boundaries.
This can help by reducing
false positive detections
in inlier pixels 
at semantic boundaries.

\subsection{Performance enhancement with negative data}
Training with negative data 
is 
an important
component of
many recent 
 outlier 
detection approaches
\cite{hendrycks19iclr,biase21cvpr,chan21iccv,grcic22eccv,tian22eccv}
due to the potential 
to address feature collapse.
In the case of dense prediction 
this usually involves 
pasting
negative data
over the 
inlier training images \cite{bevandic19gcpr,grcic21arxiv}.
Existing implementations require 
an additional loss term in negative pixels 
\cite{liu20neurips,hendrycks19iclr,chan21iccv}.
On the contrary, 
our approach does not require
any changes in the model
or the loss function.

We propose to set 
the ground truth 
of negative pixels
to \textit{void} class.
This instructs all masks
to steer clear of
 negative pixels.
This is reasonable since void pixels do not belong to any class of interest.
Such training increases
the variety of void content 
and masks get penalized
if they claim any.

The standard dense classifiers \cite{chen18eccv,orsic21pr} 
cannot be trained with negatives labeled as void.
Reason for this lies in the standard per-pixel cross-entropy loss which is not computed in void pixels.
Hence, our pasting procedure is specific for mask-level recognition.

\begin{figure}
    \centering
    \includegraphics[width=0.95\linewidth]{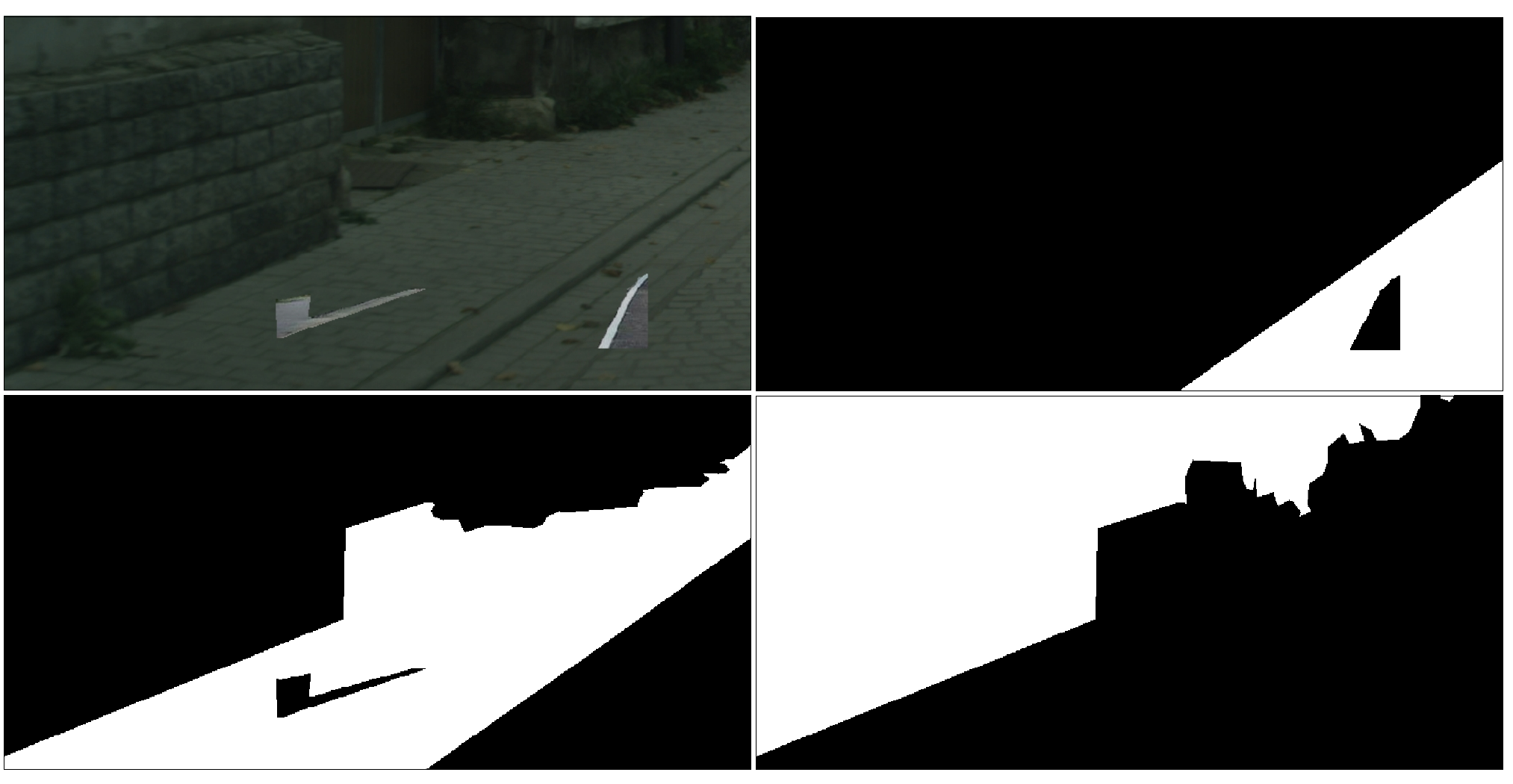}
    \caption{
        Mixed-content training image (top left)
        and mask-assignment groundtruth
        for classes road (top-right),
        sidewalk (bottom-left) and 
        building (bottom-right).
        The model is trained to reject
        the two pasted negative instances
        from all masks.
}
    \label{fig:masks}
\end{figure}

Figure \ref{fig:masks} shows a training example: the input image crop and the corresponding ground truth binary labels.
None of the ground truth labels encapsulate
the pasted negative pixels.
Our experiments show that 
this kind of supervision generalizes
to outlier detection 
in real-world images.

\section{Experiments}
\label{sec:experiments}

Our experiments explore advantages 
of mask-level recognition for 
in outlier-aware semantic segmentation.
We consider semantic (Sec.\ \ref{sec:ss_exp}) and panoptic segmentation (Sec.\ \ref{sec:ps_exp}).
\subsection{Outlier-aware segmentation of road-driving images}
\label{sec:ss_exp}
We evaluate outlier-aware 
segmentation performance 
on two standard benchmarks.
The Fishyscapes benchmark 
includes two tracks
that focus on urban road driving
\cite{blum21ijcv}.
The FS L\&F track relabels a subset 
of the Lost and Found dataset.
The FS Static track 
pastes anomalous objects
in images from Cityscapes val.
The SMIYC benchmark
(Segment Me If You Can)
includes two tracks
with real-world anomalies
in very diverse environments.
The Anomaly Track includes large anomalies
that can occur anywhere in the image,
while the Obstacle Track focuses 
on small anomalies on the road surface.

We measure the performance 
of OOD detection 
according to average precision (AP) 
and FPR at TPR of 95\%
($\mathrm{FPR}_{95}$).
We use Mask2Former (M2F)
\cite{cheng22cvpr} with
Swin-L \cite{liu21iccv} backbone.
Following the usual conventions,
we train our models
in two regimes:
with and without 
negative data.
Our models without
auxiliary data
consider only 
Cityscapes images
\cite{cordts16cvpr}.
This likely reduces 
our performance
on SMIYC due to 
large domain shift
\cite{vaze22iclr}. 
Models with negative
data are first trained 
with Cityscapes taxonomy
on images from Cityscapes
and Mapillary Vistas \cite{neuhold17iccv}.
Then, we fine-tune
the model
for 2K iterations
on mixed-content images.
We paste ADE20K \cite{zhou19ijcv} instances
as negative data.
We use standard hyper-parameters
\cite{cheng22cvpr}
except for the batch size,
which we set to 18.
The longest experiments 
last about 48 hours 
on 3$\times$A6000.

Table \ref{tbl:smiyc} 
compares the performance 
of our best approach (M2F-EAM)
with the related work on SMIYC.
The two sections organize the methods
depending on whether they train 
on real negative data.
Our model trained
without negative data
achieves strong 
average precision 
in both tracks.
High AP and comparatively poor
$\mathrm{FPR}_{95}$ scores
suggest rare occurrences
of highly confident
false negative detections.
Analysis of the AUROC curve 
supports this hypothesis 
since we achieve
$\mathrm{FPR}_{90}$ = $20\%$.

Training on more diverse 
closed-set images
and fine-tuning with 
negative data
significantly 
improves the results.
Moreover, our model 
trained with auxiliary data
achieves state-of-the-art 
performance on SMIYC benchmark 
across all metrics.
Dramatic improvement in FPR 
suggests that
training with negative data
improves models ability
to detect diverse anomalies.

\begin{table}[htb]
\begin{footnotesize}
\begin{tabular}{lccccc}
\multirow{2}{*}{Method} & \multicolumn{1}{c}{\multirow{2}{*}{Aux}} & \multicolumn{2}{c}{AnomalyTrack} & \multicolumn{2}{c}{ObstacleTrack} \\ 
 & \multicolumn{1}{c}{data} & \multicolumn{1}{c}{AP} & \multicolumn{1}{c}{$\mathrm{FPR}_{95}$} & AP & \multicolumn{1}{c}{$\mathrm{FPR}_{95}$}  \\ \hline
Image Resyn. \cite{lis19iccv} & \xmark& 52.3  & \textbf{25.9} & 37.7 & \textbf{4.7}\\ 
Road Inpaint. \cite{lis20arxiv} & \xmark& - & -  & 54.1 & 47.1  \\ 
JSRNet \cite{vojir21iccv} & \xmark & 33.6& 43.9 & 28.1 & 28.9 \\ 
Max softmax \cite{hendrycks17iclr} & \xmark & 28.0 & 72.1 & 15.7 & 16.6  \\ 
MC Dropout \cite{kendall17nips} & \xmark & 28.9 & 69.5 & 4.9 & 50.3 \\ 
ODIN \cite{liang18iclr} & \xmark & 33.1 & 71.7 & 22.1 & 15.3  \\ 
Embed.\ Dens.\ \cite{blum21ijcv} & \xmark & 37.5 & 70.8 & 0.8 & 46.4 
\\[0.3em]
M2F-EAM (ours) & \xmark & \textbf{76.3} & 93.9 & \textbf{66.9} & 17.9 \\
\hdashline
SynBoost \cite{biase21cvpr} & \cmark& 56.4 & 61.9 & 71.3 & 3.2 \\ 
DenseHybrid \cite{grcic22eccv} & \cmark& 78.0 & 9.8  & 78.7 & 2.1 \\ 
PEBAL \cite{tian22eccv} & \cmark&  49.1 & 40.8 & 5.0 & 12.7 \\ 
Void Classifier \cite{blum21ijcv} & \cmark & 36.6 & 63.5 & 10.4 & 41.5 \\
Maxim. Ent. \cite{chan21iccv} & \cmark & {85.5}  & {15.0} & {85.1} & {0.8} \\[0.3em]
M2F-EAM (ours) & \cmark & \textbf{93.8} & \textbf{4.1} & \textbf{92.9} & \textbf{0.5} \\
\end{tabular}
\end{footnotesize}
\centering
\caption{Outlier-aware segmentation on SMIYC. 
 Our AP performance outperforms 
 all previous approaches 
 in both categories.
}
\label{tbl:smiyc}
\end{table}

Table~\ref{tab:fs_test} compares
our method (M2F-EAM) with related work
on the Fishyscapes benchmark \cite{blum19iccvw}.
As before,
the two sections 
gather methods 
based on whether 
they use real 
negative data (bottom)
or not (top).
Our method achieves
the best performance 
on FS Static
in both categories and 
the best AP performance 
on FS Lost and Found.

\begin{table}[h]
\begin{footnotesize}
\begin{tabular}{lccccc}
\multirow{2}{*}{Method} &  \multicolumn{2}{c}{FS L\&F} & \multicolumn{2}{c}{FS Static} & CS Val \\
  & AP & $\mathrm{FPR}$  & AP & $\mathrm{FPR}$ & mIoU \\ \hline
{Maxim. Ent. \cite{chan21iccv}}  & {15.0} & 85.1 & 0.8 & 77.9 & 9.7 \\
Image Resyn. \cite{lis19iccv}  & 5.7 & 48.1 & 29.6 & 27.1 & 81.4  \\
Max softmax \cite{hendrycks17iclr}  &  1.8 & 44.9 & 12.9 & 39.8 & 80.3 \\
SML \cite{jung21iccv}  & 31.7 & 21.9 & 52.1 & 20.5 & -\\
Embed.\ Dens.\ \cite{blum21ijcv}  & 4.3 & 47.2 & 62.1 & 17.4 & 80.3\\
NFlowJS \cite{grcic21visapp}  & 39.4 & 9.0 & 52.1 & 15.4 & 77.4 \\
SynDHybrid \cite{grcic22eccv} & {51.8} & {11.5} & 54.7 & {15.5} & 79.9 \\[0.2em]
M2F-EAM (ours) & 9.4 & 41.5 & \textbf{76.0} & \textbf{10.1} & 83.5 \\
\hdashline
SynBoost \cite{biase21cvpr}  & 43.2 & 15.8 & 72.6 & 18.8 & 81.4 \\
Prior Entropy \cite{malinin18nips}  & 34.3 & 47.4 & 31.3 & 84.6 & 70.5 \\
OOD Head \cite{bevandic22ivc}  & 30.9 & 22.2 & 84.0 & 10.3 & 77.3 \\
Void Classifier \cite{blum21ijcv}  & 10.3 & 22.1 & 45.0 & 19.4 & 70.4 \\
Dirichlet prior \cite{malinin18nips}  &  34.3 & 47.4 & 84.6 & 30.0 & 70.5\\
DenseHybrid \cite{grcic22eccv}  & {43.9} & \textbf{6.2} & 72.3 & {5.5} & {81.0} \\
PEBAL \cite{tian22eccv} & 44.2 & 7.6 & 92.4 & 1.7 & - \\[0.2em]
M2F-EAM (ours) & \textbf{63.5} & 39.2 & \textbf{93.6} & \textbf{1.2} & {83.5} \\
\end{tabular}
\end{footnotesize}
\centering
\caption{Outlier-aware segmentation on Fishyscapes benchmark. 
 Our AP performance outperforms 
 all previous approaches.
}
\label{tab:fs_test}
\end{table}

Table \ref{tab:fs_val} evaluates
outlier-aware segmentation
on validation subsets 
of Road Anomaly \cite{lis19iccv} 
and Fishyscapes \cite{blum19iccvw}.
We compare our 
mask-level approaches
with the standard pixel-level baseline
(PerPixel)
and the previous work.
Again, methods from the bottom section
train on auxiliary negative data
while the others see only inliers.
Our two mask-level approaches
outperform the pixel-level baseline
and all previous approaches.
Among the two 
mask-level approaches,
ensemble over anomaly scores 
(M2F-EAM) outperforms
anomaly score of the ensemble
(M2F-AEM).

\begin{table}[ht]
\centering
\begin{footnotesize}
\begin{tabular}{lccccccc}
\multirow{2}{*}{Model} & \multicolumn{2}{c}{Road Anomaly} & \multicolumn{2}{c}{FS L\&F} & \multicolumn{2}{c}{FS Static} \\
  & AP & $\mathrm{FPR}$  & AP & $\mathrm{FPR}$ & AP & $\mathrm{FPR}$ \\ \hline

MSP \cite{hendrycks17iclr} & 15.7 & 71.4 & 4.6 & 40.6 & 19.1 & 24.0\\
ML \cite{hendrycks22icml} & 19.0 & 70.5 & 14.6 & 42.2 & 38.6 & 18.3\\
NFlowJS \cite{grcic21arxiv} &  - & - & 40.2 & 18.7 & 34.4 & 11.2 \\
SML \cite{jung21iccv} & 25.8 & 49.7 & 36.6 & 14.5 & 48.7 & 16.8 \\
SynthCP \cite{xia20eccv} & 24.9 & 64.7 & 6.5 & 46.0 & 23.2 & 34.0 \\
Density \cite{blum21ijcv} &  - & - & 4.1 & 22.3 & - & - \\[0.3em]
PerPixel  & 49.3 & 31.0 & 2.5 & 56.7 & 11.5 & 34.8  \\
M2F-AEM  & \textbf{66.9} & 15.3 & 51.2 & 28.0 & 86.2 & 3.5 \\
M2F-EAM  & \textbf{66.7} & \textbf{13.4} & \textbf{52.0} & \textbf{20.5} & \textbf{87.3} & \textbf{2.1} \\


\hdashline

SynBoost\cite{biase21cvpr} & 38.2 & 64.8 & 60.6 & 31.0 & 66.4 & 25.6\\
Energy \cite{liu20neurips} & 19.5 & 70.2 & 16.1 & 41.8 & 41.7 & 17.8 \\
PEBAL\cite{tian22eccv} & 45.1 & 44.6 & 58.8 & 4.8 & 92.1 & 1.5\\
DenseHybrid \cite{grcic22eccv}& - & - & 63.8 & 6.1 & 60.0 & 4.9\\[0.2EM]
M2F-EAM  & \textbf{69.4} & \textbf{7.7} & \textbf{81.5} & \textbf{4.2} & \textbf{96.0} & \textbf{0.3}\\

\end{tabular}
\end{footnotesize}
 \caption{Comparison of our 
   mask-level approaches
   (M2F-EAM, M2F-AEM)
   with the pixel-level baseline
   (PerPixel) and the previous work
   on RoadAnomaly and Fishyscapes val.
 }
\label{tab:fs_val}
\end{table}

\subsection{Outlier-aware panoptic segmentation on MS COCO}
\label{sec:ps_exp}

Mask-level outlier detection 
can also be applied for panoptic segmentation.
We consider the hardest setup 
from a recent related work 
\cite{hwang21cvpr} 
that relabels 20\% 
of thing classes from COCO
as void pixels during training.
These classes are dining table,
banana, bicycle, cake, sink, cat,
keyboard, and bear.
During inference the model has 
to classify all pixels 
from these classes 
into the dedicated 
anomalous thing class.
Outlier-aware performance
is measured according to
standard metrics PQ, SQ, and RQ.
Our models use 
a ResNet-50 backbone 
as in the previous work
\cite{hwang21cvpr}.

Mask-level training 
encourages all masks to refrain 
from encompassing the void pixels.
Our anomaly detectors 
are sensitive to the resulting 
lack of mask assignment.
Hence, the intensity 
of our supervision 
is very similar 
to void-suppression
\cite{hwang21cvpr}.
Our inference recovers
the dense anomaly map
by thresholding the
mask-level anomaly score.
We validate the threshold
for 95\% TPR in outlier detection
on a held-out validation image.
We assign each anomalous pixel
to its prefered mask 
and form instances by 
keeping all masks
with more than 200 pixels.

Table \ref{tab:osp-coco} 
compares our method 
to several approaches 
from the EOPSN paper
\cite{hwang21cvpr}.
We outperform 
all previous work,
in spite of much less supervision.
Note that our method
can easily accommodate 
anomalous stuff classes.

\begin{table}[ht]
\centering
\begin{footnotesize}
\begin{tabular}{lcccccc}
  \multirow{2}{*}{Method} & \multicolumn{3}{c}{Known} & \multicolumn{3}{c}{Unknown} \\
 & PQ & SQ & RQ & PQ & SQ & RQ \\ \hline
 Void-background & 37.7 & 76.3 & 46.6 & 4.0 & 71.1 & 5.7 \\
 Void-ignorance & 37.2 & 76.3 & 45.9 &  3.7 & 71.8 & 5.2 \\
 Void-suppression & 37.5 & 75.9 & 46.1 & 7.2 & \textbf{75.3} & 9.6 \\
 Void-train & 36.9 & 76.4 & 45.5 & 7.8 & 73.4 & 10.7 \\
 EOPSN \cite{hwang21cvpr} & 37.4 & 76.2 & 46.2 & 11.3 & 73.8 & 15.3 
 \\[0.3em]
  Open-M2F-AEM & \textbf{43.5} & \textbf{82.0} & \textbf{52.2}  & 11.3 & 73.3 & 15.3 \\
 Open-M2F-EAM & \textbf{43.5} & \textbf{82.0} & \textbf{52.2}  & \textbf{13.2} & 73.4 & \textbf{18.0} \\
\end{tabular}
\end{footnotesize}
\caption{outlier-aware 
  panoptic segmentation on COCO.
  We relabel 20\% of thing classes
  to the unknown void class
  \cite{hwang21cvpr}.
  We outperform other approaches
  both on known and unknown classes.
  }
\label{tab:osp-coco}
\end{table}

\begin{figure}[ht]
  \centering
  \includegraphics[width=\linewidth]
  {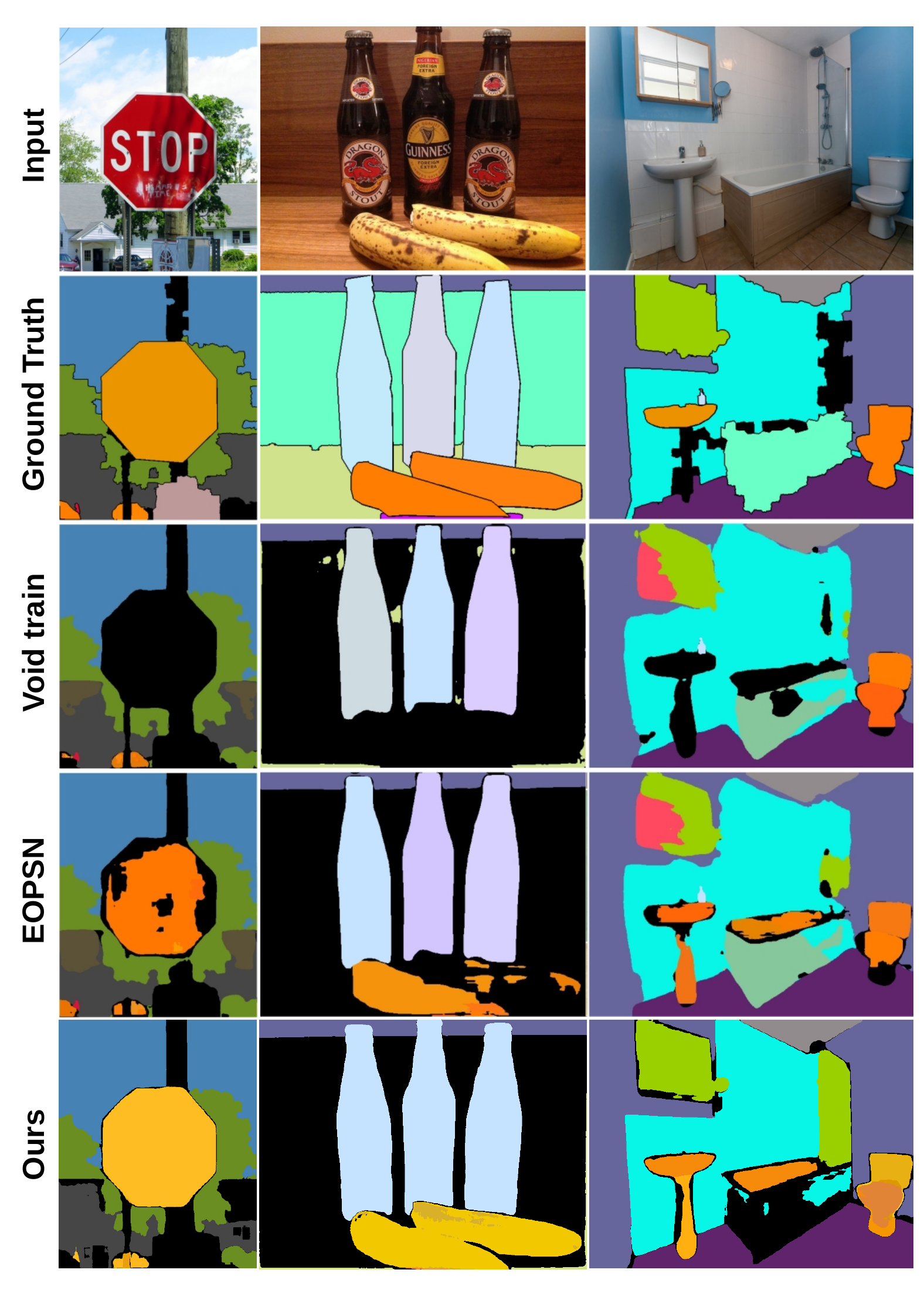}
  \caption{
   outlier-aware panoptics
   with M2F-EAM.
   Stop sign, bananas, toilet 
   and sink are considered 
   unknown thing classes
   \cite{hwang21cvpr}.
   We detect all unknown classes
   and distinguish some instances.
  }
  \label{fig:panoptic_fig}
\end{figure}

Figure \ref{fig:panoptic_fig} 
shows qualitative results
on three scenes from COCO val.
The rows show:
input image,
ground truth,
two results from \cite{hwang21cvpr}
and finally our results.
The results 
clearly illustrate improvements
of our method 
over previous state of the art 
in outlier-aware panoptic segmentation.

Note finally that 
panoptic mask-level models 
can also be used for 
standard outlier-aware semantic segmentation.
In fact, panoptic models outperform
their semantic counterparts 
in 3 out of 6 metrics
from Table \ref{tab:fs_val}.

\section{Ablations}
\label{sec:ablations}

We ablate the choice of the OOD score (Sec.\ \ref{sec:abl_oods}), the backbone (Sec.\ \ref{sec:abl_bb}), the number of masks (Sec.\ \ref{sec:abl_mc}), and the source of negative data (sec.\ \ref{sec:abl_ds}).

\subsection{Impact of the OOD score}
\label{sec:abl_oods}

Table \ref{tab:pixel_vs_mask} 
considers several OOD detectors 
that can be plugged into 
our methods.
The five sections consider 
per-pixel baseline and the aforementioned M2F-AM, M2F-AHM, M2F-AEM, and M2F-EAM.
We note that neither 
ensembles of mask scores nor 
the mask scores themselves
are distributions.
Hence we do not consider
probabilistic anomaly detectors
in the last four sections.
Instead, we only consider
simply taking the hard maximum
(this is related to max-softmax)
or the energy score (log-sum-exp).
The two options perform comparably
so we choose to use hard maximum
in our submissions to SMIYC
as a simpler choice.
As before,
we observe slight advantage
of M2F-EAM over M2F-AEM,
as well as poor performance
of per-pixel outlier detection
that is in line with previous work
\cite{blum21ijcv,hendrycks22icml}.
Additionally,
we observe that 
ensemble-based methods 
outperform their simpler 
counterparts M2F-AM and M2F-AHM.
\begin{table}[h]
\centering
\begin{footnotesize}
\begin{tabular}{llcc}
Method & Anomaly detector & FS L\&F & FS Static \\\hline
\multirow{4}{*}{PerPixel} & Entropy \cite{hendrycks19iclr}  & 2.9   & 12.7 \\ 
 & KL div \cite{grcic21visapp}  & 4.1  & 16.4\\ 
 & Energy \cite{liu20neurips}  & 2.4  & 11.3\\ 
 & Max-softmax \cite{hendrycks17iclr}   & 1.8 & 8.9 \\ \\

M2F-AM
& Max-score & 30.9 & 30.2 \\\\

M2F-AHM
& Max-score & 3.5 & 44.4 \\\\
 
\multirow{2}{*}{M2F-AEM} 
& Energy  & 51.1 & 86.6 \\
& Max-score  & 51.2 & 86.2 \\\\

\multirow{2}{*}{M2F-EAM} 
 & Energy   & 48.5  & 69.3 \\ 
 & Max-score & \textbf{52.0}  & \textbf{87.3}
\end{tabular}
\end{footnotesize}
 \caption{Validation of 
  anomaly detectors 
  that can plug-in
  into our methods.
  Energy score (log-sum-exp)
  performs similar to 
  taking a hard maximum.
  Again, M2F-EAM outperforms
  M2F-AEM while both 
  mask-level approaches
  outperform M2F-AM, M2F-AHM, and 
  per-pixel baseline.
 }
\label{tab:pixel_vs_mask}
\end{table}

\subsection{Impact of the backbone}
\label{sec:abl_bb}

Table \ref{tab:fishyval} 
investigates OOD 
detection performance
of per-pixel and mask-classification
models with different backbones.
We consider two convolutional backbones,
ResNet-50 and a more advanced ConvNeXt-L.
We also consider transformer-based 
backbone Swin-L.
Additionally, we show results
of DeepLabV3+ model with 
ResNet-50 backbone.
Our per-pixel baseline 
and DLv3+ perform similarly while Mask2Former 
outperforms both methods.
Strong performance 
of M2F models based on Swin-L
suggests that large capacity
and transformer architecture
may be important for
mask-based outlier-aware segmentation.
\begin{table}[h]
\centering
\begin{footnotesize}
\begin{tabular}{llc@{\quad}cc@{\quad}cc}
\multirow{2}{*}{Backbone} & \multirow{2}{*}{Model} & \multicolumn{2}{c}{FS L\&F} & \multicolumn{2}{c}{FS Static} & CS val \\
 &  & AP & $\mathrm{FPR}$ & AP & $\mathrm{FPR}$ & mIoU \\ \hline
\multirow{3}{*}{ResNet-50} & DLv3+ & 3.5 & 45.0 & - & - & 77.8 \\
 & PerPixel & 1.3 & 64.0 & 9.0 & 42.9 & 79.6 \\
 & M2F-EAM & 20.8 & 22.7 & 36.7 & 23.8 & 79.4 \\ \\

\multirow{1}{*}{ConvNeXt-L} & M2F-EAM & 31.5 & 28.6 & 76.3 & 6.3 & 82.6 \\\\
  
\multirow{2}{*}{Swin-L} & PerPixel & 2.5 & 56.7 & 11.5 & 34.8 & 83.2 \\
  & M2F-EAM  & \textbf{52.0} & \textbf{20.5} & \textbf{87.3} & \textbf{2.1} & \textbf{83.5} \\
\end{tabular}
\end{footnotesize}
\caption{Comparison of 
  several models
  with different backbones  
  on Fishyscapes val.
  Mask-level models
  outperform their 
  per-pixel counterparts,
  and this is a major 
  takeaway of our work.
}
\label{tab:fishyval}
\end{table}

\subsection{Impact of the mask count}
\label{sec:abl_mc}

Table \ref{tab:mask_count} 
explores the significance
of the number of masks N
for closed-set recognition 
and outlier detection.
We consider the case where 
the number of masks equals 
the number of classes (N=19)
as well as two 
more abundant choices (N=50,100).
These experiments reveal
a very strong influence of N
to outlier detection performance,
although both tasks profit
from having many masks.
\begin{table}[ht]
\centering
\begin{footnotesize}
\begin{tabular}{cccccc}
\multirow{2}{*}{Mask count} & \multicolumn{2}{c}{FS L\&F} & \multicolumn{2}{c}{FS Static} & CS val \\
 & AP & $\mathrm{FPR}_{95}$ & AP & $\mathrm{FPR}_{95}$ & mIoU \\ \hline
19 & 33.5 & 18.7 &72.5 & 6.8 & 82.8\\
 50 & 47.9 & 24.7 & 69.7 & 4.8 & 83.1\\
 100 & \textbf{52.0} & \textbf{20.5} & \textbf{87.3} & \textbf{2.1} & \textbf{83.5} \\
\end{tabular}
\end{footnotesize}
  \caption{Impact of mask count 
    to outlier detection 
    and closed-set segmentation
    with M2F-EAM. 
    Abundant set of masks 
     improves resilience to outliers.
  }
  \label{tab:mask_count}
\end{table}

\subsection{Impact of the negative data source}
\label{sec:abl_ds}

Table~\ref{tab:fs_neg_val}
validates different
sources of negative data
on validation subsets
of Road Anomaly and Fishyscapes.
The first row shows 
the results without 
negative data training.
The second row 
corresponds to
pasting randomly selected
square patches
from other images
of the batch
atop the considered image.
The third row
corresponds to 
pasting patches
generated by a
normalizing flow model
trained only on the
inlier images.
The last row
corresponds to pasting 
instances from ADE20K,
cut according
to their GT mask.
The results show
that pasting 
ADE20K instances
outperforms all approaches
on Fishyscapes.
It achieves the best FPR
and comparable AP score
on Road Anomaly.
Thus, we chose this
as our default setup
when training with 
negative data.

\begin{table}[ht]
\centering
\begin{footnotesize}
\begin{tabular}{lc@{\ }cc@{\ }cc@{\ }c}
\multirow{2}{*}{Negatives} & \multicolumn{2}{c}{Road Anom.} & \multicolumn{2}{c}{FS L\&F} & \multicolumn{2}{c}{FS Static} \\
  & AP & $\mathrm{FPR}$ & AP & $\mathrm{FPR}$ & AP & $\mathrm{FPR}$ \\ \hline
  w/o negatives & 66.7 & 13.4  & 52.0 & 20.5 & 87.3 & 2.1 \\
Inlier patches  & \textbf{69.7} & 8.8 & 77.0 & 10.1 & 95.8 & 0.7 \\
Generated samples  & 68.9 & 8.4 & 80.6 & 4.5 & 91.9 & 0.9 \\
ADE20K instances  & 69.4 & \textbf{7.7} & \textbf{81.5} & \textbf{4.2} & \textbf{96.0} & \textbf{0.3}\\
\end{tabular}
\end{footnotesize}
\vspace{-0.2cm}
 \caption{Validation of various kinds of negative data.
 Broad negative dataset outperforms other alternatives.}
\label{tab:fs_neg_val}
\end{table}

\newpage
\section{Conclusion}
\label{sec:conclusion}

Robust performance in presence of outliers 
is an important prerequisite
for many exciting applications
of scene understanding.
Most previous dense prediction approaches 
build on pixel-level OOD detection
and thus fail to account for the correlation 
between neighbouring pixels.
We address this research problem
by shifting OOD detection from pixels to regions.
The resulting mask-level predictions
aggregate pixel-level evidence
and thus increase 
the statistical power 
of the corresponding anomaly scores.
We also show that
it is especially beneficial
to perform OOD detection
before ensembling decisions
over particular masks.
We further boost our performance by injecting
negative data into void content.
Finally, we extend mask-based model
for panoptic inference in the presence of outliers.
Experiments reveal that
mask-level outlier detection
outperforms 
pixel-level counterparts 
by a wide margin 
and achieves 
state-of-the-art AP performance 
among methods that do not train 
on real negative data.
Furthermore, it also improves upon
the previous state of the art
in outlier-aware panoptic segmentation
in spite of requiring  less supervision
than previous work.
The proposed formulation of 
mask-level outlier-aware segmentation 
can accommodate any anomaly detector
based on discriminative 
recognition score,
and can be combined 
with many previous approaches.
Promising directions for future work
include learning 
with synthetic negatives
and modelling probabilistic density
of mask-wide descriptors.
The source code will be 
available upon publication.

\section{Limitations}

In spite of accomplishing 
very competitive AP scores,
our approach may produce 
poor FPR95 performance if 
an outlier object
resembles a known class. 
Still, this can be successfully alleviated with negative training data as shown in the experiments.


\section*{Acknowledgements}
This work has been supported by Croatian Science Foundation grant IP-2020-02-5851 ADEPT, by NVIDIA Academic Hardware Grant Program, as well as by European Regional Development Fund grants KK.01.1.1.01.0009 DATACROSS and KK.01.2.1.02.0119 A-Unit.

\clearpage
{
    \small
    \bibliographystyle{ieee_fullname}
    \bibliography{references}
}

\end{document}